\documentclass[conference]{IEEEtran}
\IEEEoverridecommandlockouts
\usepackage{cite}
\usepackage{amsmath,amssymb,amsfonts}
\usepackage{algorithmic}
\usepackage{graphicx}
\usepackage{textcomp}
\usepackage{xcolor}
\usepackage{booktabs}
\usepackage{siunitx}

\def\BibTeX{{\rm B\kern-.05em{\sc i\kern-.025em b}\kern-.08em
    T\kern-.1667em\lower.7ex\hbox{E}\kern-.125emX}}
\begin{document}

\title{Facial Features Integration in Last Mile Delivery Robots\\

}

\author{\IEEEauthorblockN{Delgermaa Gankhuyag}
\IEEEauthorblockA{\textit{Department Intelligent Transport Systems} \\
\textit{Johannes Kepler University}\\
Linz, Austria \\
delgermaa.gankhuyag@jku.at}
\and
\IEEEauthorblockN{Stephanie Groi{\ss}}
\IEEEauthorblockA{\textit{} 
\textit{Johannes Kepler University}\\
Linz, Austria\\
k00655913@students.jku.at}
\and
\IEEEauthorblockN{Lena Schwamberger}
\IEEEauthorblockA{\textit{} 
\textit{Johannes Kepler University}\\
Linz, Austria \\
k01320196@students.jku.at}
\and
\IEEEauthorblockN{\hspace{2cm}{\"O}zge Talay}
\IEEEauthorblockA{\textit{} 
\hspace{2cm}\textit{Johannes Kepler University}\\
\hspace{2cm}Linz, Austria \\
\hspace{2cm}k11712793@students.jku.at}
\and
\IEEEauthorblockN{\hspace{-2cm}Cristina Olaverri-Monreal}
\IEEEauthorblockA{\hspace{-2cm}\textit{Department Intelligent Transport Systems} \\
\textit{\hspace{-2cm}Johannes Kepler University}\\
\hspace{-2cm}Linz, Austria \\
\hspace{-2cm}Cristina.Olaverri-Monreal@jku.at}

}

\maketitle

\begin{abstract}
Delivery services have undergone technological advancements, with robots now directly delivering packages to recipients. While these robots are designed for efficient functionality, they have not been specifically designed for interactions with humans. Building on the premise that incorporating human-like characteristics into a robot has the potential to positively impact technology acceptance, this study explores human reactions to a robot characterized with facial expressions. The findings indicate a correlation between anthropomorphic features and the observed responses.
\end{abstract}


\section{Introduction}
The rapid expansion of e-commerce and urbanization has led to a notable increase in urban freight logistics, particularly in the last-mile delivery (LMD) phase from suppliers to customers. Despite its crucial role in ensuring customer satisfaction, last-mile delivery faces challenges such as lack of automation, inconsistent and congested routes, and growing demand for same-day delivery. Both academia and industry have focused on developing sustainable, efficient, and cost-effective parcel delivery solutions. LMD robots, already deployed in several cities and projected to reach one million by 2025, offer a promising solution. Yet challenges persist according to recent research that relate to customer acceptance of LMDs and their satisfaction with delivery.  Incorporating human-like characteristics could influence human responses and potentially enhance acceptance of LMD robots.

According to~\cite{willis2006first} individuals have the ability to detect particular characteristics of others within a tenth of a second, underscoring this the significance of initial impressions. In this context, the presence of even subtle facial features has the potential to generate a response~\cite{zebrowitz1996social}, serving an extended exposure only to increase the confidence of the initial impression.
Throughout this process, the observer derives information from the counterpart's face, including details such as gender and age, forming a perceptual bias that becomes a tool for anticipating and preparing in case of further interaction (e.g. conversation).

When encountering a non-human object with similar characteristics, the same process is automatically initiated~\cite{hegel2014bedeutung}. This suggests that in this scenario, a rapid evaluation, comparable to an initial impression, takes place within an exceptionally brief period.

This paper aims to investigate the initial responses of individuals to a last-mile delivery (LMD) robot incorporating anthropomorphic characteristics. Specifically, we explored whether distinct facial expressions could serve as a foundation for improving technology acceptance and enhancing user experience. This investigation represents a contribution to the body of knowledge as such an experiment has not been conducted before.
To this end we formulated the following hypotheses: 
\begin{itemize}
\item H0: LMD robots, featuring expressive facial features, generate no response from pedestrians.
\item H1: Pedestrians exhibit reaction to LMD robots featuring expressive facial features.
\end{itemize}

The following section presents a selection of findings from studies on anthropomorphism, social robots and facial expressions. Section \ref{sec:experiment} presents the experimental setup. The model design and analyzing method is described in Section \ref{sec:variables}. The results of the conducted experiment are presented in Section \ref{sec:results}. Lastly, Section \ref{sec:conclusion} summarizes the key findings and outlines potential future directions.

\section{Background Information}
\label{sec:related}

This section presents various findings from studies that contribute to supporting the experiment design and setup.

Prior studies have shown that people have a strong inclination to actively seek out faces~\cite{song2021facial}, underscoring the crucial role of facial features in conveying social cues.


In this context, related literature defines anthropomorphism as the capacity of the human mind to attribute both physical and mental human features to non-human objects~\cite{epley2008when}. This happens because humans naturally try to understand both other humans and non-human things in the same way. In this context, the phenomenon known as pareidolia, occurs when individuals attribute human-like qualities, such as emotions or expressions, to non-human entities based on familiar patterns~\cite{sagan1995demonhaunted}. For example, a pattern that reminds to eyes and a curved line for a mouth on an  object might lead someone to perceive it as having a smile. 

Prior research showed humans' natural tendency to involuntarily anthropomorphize robots by attributing human-like traits to them~\cite{hegel2014bedeutung}. Nevertheless, when robots become excessively anthropomorphic, they can be perceived as potentially threatening. 

According to~\cite{zlotowski2014anthropomorphism} and ~\cite{mori2012uncanny}, there exists a correlation between favorability towards a robot and its degree of resemblance to a human, up to a specific threshold. Beyond this similarity threshold, there is a decrease in likability, leading to a tendency to avoid the robot. Even atypical human features, such as enlarged eyes or mismatched facial features, can trigger feelings of discomfort~\cite{mende2019use}. 


A customer's decision to purchase a product can be influenced by integrating human-like characteristics into the product~\cite{aggarwal2007car}. Visual stimuli, such as facial features, play a vital role in supporting customers during the decision-making process. Studies suggest that the facial design significantly impacts whether a product is bought, as the judgment of potential buyers are influenced, leading them to favor anthropomorphic products~\cite{guthrie1993faces}.   

Similarly,  optimizing the anthropomorphic features of a LMD robot could potentially enhance its appeal and overall acceptance among customers, and other actors that might encounter them.

In this context the need for a thoughtful design rather than just a simple inclusion of features is crucial~\cite{steffen2009brain}.


Building upon prior research that showed humans' natural tendency to involuntarily anthropomorphize robots by attributing human-like traits to them ~\cite{hegel2014bedeutung}, this paper aims to enhance the capabilities of a mobile LMD robot. Specifically, we contribute to the state of the art by proposing the integration of a representation of facial features capable of depicting two distinct emotions, potentially advancing fields such as emotion recognition and human-computer interaction.

\section{Experimental Setup}
\label{sec:experiment}

\subsection{Apparatus}
The LMD robot from the Intelligent Transport System Department at the Johannes Kepler University Linz (JKU-ITS) is equipped with various sensors (3D LIDAR, RGB-D camera, IMU, and GPS) and a computer for processing sensor input and obstacle detection~\cite{novotny2023ros}. For safety during the study, the robot was not autonomously operated but remotely controlled by members of the JKU-ITS team.

\begin{figure}[!t]
	\centering
	\includegraphics[width=0.5\textwidth]{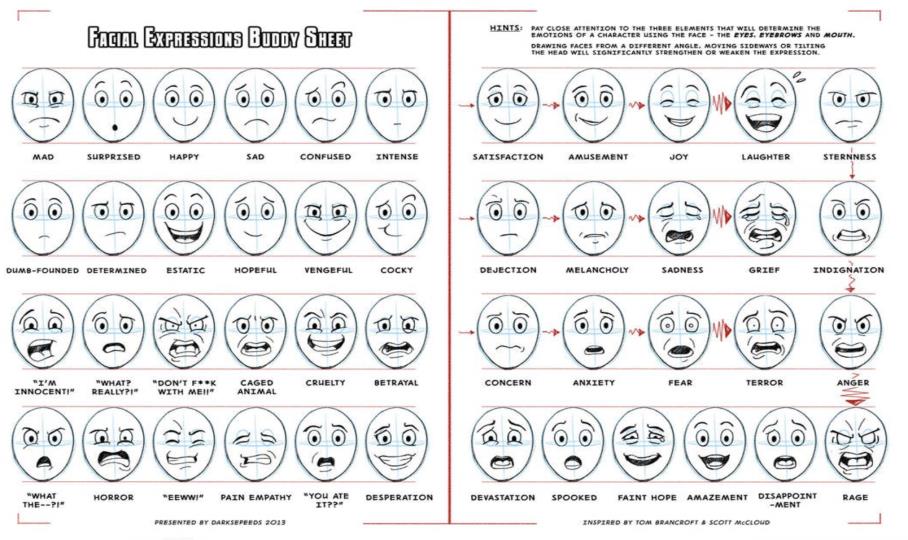}
	\caption{Facial expressions design sheet utilized to select the optimal facial feature for the conducted experiment~\cite{darkspeeds}}
	\label{fig:facial}
\end{figure}

The facial features selected for the experiment in this study represented happy and angry emotions, as positive and negative emotions are typically recognized faster than neutral expressions in people~\cite{steffen2009brain}. The happy expression was characterized by arched eyes and a smiling mouth, while the angry expression displayed lowered eyebrows narrowed eyes and downturned mouth. These facial expressions relied on a facial expressions design sheet commonly used by artists as shown in Fig.~\ref{fig:facial}.

\begin{figure}[!t]
	\centering
	\includegraphics[width=0.3\textwidth]{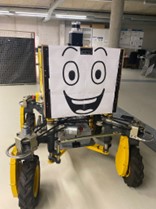}
	\caption{LMD robot with the happy face.}
	\label{fig:LMDRobotHappy}
\end{figure}

\begin{figure}[!t]
	\centering
	\includegraphics[width=0.3\textwidth]{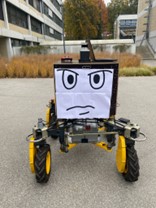}
	\caption{LMD robot with the angry face.}
	\label{fig:LMDRobotAngry}
\end{figure}

After selecting the ideal expressions corresponding to happiness and anger, they were resized to fit the front side of the robot for a greater visual impact (see Fig. \ref{fig:LMDRobotHappy} and Fig.~\ref{fig:LMDRobotAngry}).

\begin{figure*}[!t]
	\centering
	\includegraphics[width=0.8\textwidth]{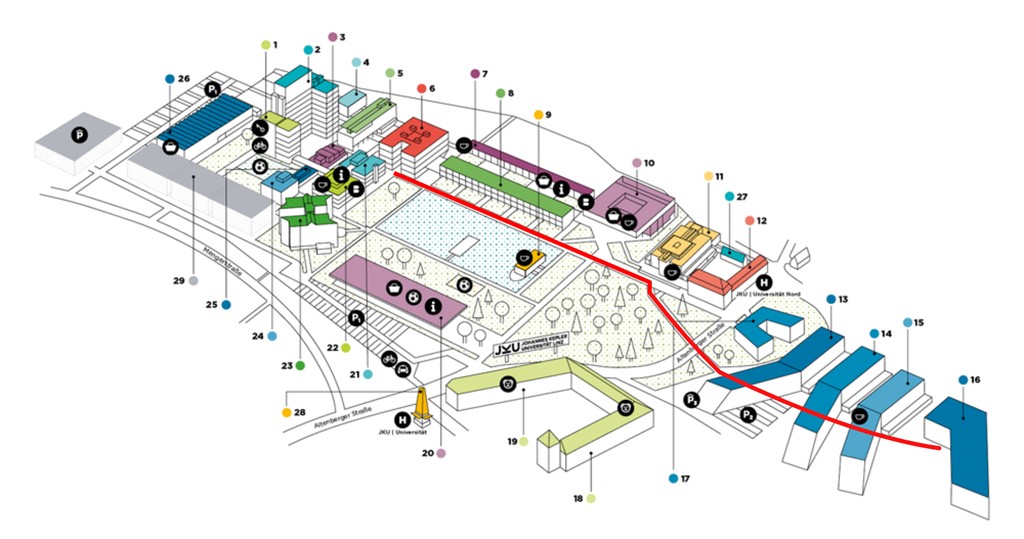}
	\caption{Adapted figure representing the Johannes Kepler University Campus~\cite{jku}}
	\label{fig:campus}
\end{figure*}

\subsection{Conducted Experiment}
We conducted distinct iterations of the experiment to gather ample observations and compare a baseline condition, with no facial expressions, against expressions conveying anger or happiness.
The experiment was performed during the lunch break, from 11 am to 1 pm, along the pathway from the JKU campus science park area to the Juridicum building, as indicated by the red line in Fig. \ref{fig:campus}. 
This specific location and time were selected due to the considerable pedestrian activity. Permission to conduct the experiment on campus grounds was secured in advance.

Pedestrians' responses to the robot were documented during the experiment, leading to the creation of three distinct categories:

\begin{itemize}
\item positive reaction: people who showed a smile
\item negative reaction: those displaying discomfort  
\item neutral reaction: all other facial expressions are included in this category
\end{itemize}

In addition, three specific types of interpersonal engagement were distinguished: 
\begin{itemize}
\item  Eye contact: defined as direct visual engagement with the robot. 
\item Experimenting: included behaviors such as blocking the robot's path, verbal communication, and taking pictures or videos.
\item Stopping: was defined as a person that stopped during the interaction.\\
\end{itemize}

\section{Variables selection, model design and analyzing methods}
\label{sec:variables}

A Cramér's V and Chi-Square statistical analysis were performed to clarify relationships among the variables of interest, followed by fitting a logit model to the dataset.

To investigate pedestrians' responses to the LMD robot, we preselected the following variables prior to the data collection phase.
  
\subsection{Response variable}
To understand pedestrians' responses to the LMD robot exhibiting various facial expressions, we selected the pedestrians' reaction variable (referred to as ``Emotion'' in the modeling) with three distinct levels: positive, negative, and neutral emotions.

\subsection{Covariates}
We assumed that several variables would influence pedestrians' reactions. Therefore, the following factors were selected as covariates for the modeling:
\begin{itemize}
\item Interaction: This variable indicates whether there was an interaction between pedestrians and the robot, categorized as ``Yes'' or ``No''.
\item Gender: Categorized into ``Male'' and ``Female''.
\item  RobotFace: Different facial expressions including ``HappyFace'', ``AngryFace'', and ``NeutralFace'', representing treatment and control scenarios in the experiment.
\item Age: Categorized as ``Young'', ``Middle'', and ``Senior''.
\item Contact: Categorized into ``Eye contact'', ``Experimenting'', ``Stopping'', and ``None'', representing the level of interpersonal engagement of pedestrians as described in Section~\ref{sec:experiment}.
\end{itemize}

\subsection{Model design}
We applied a binary logistic regression model to the categorical response variable, Emotion. Below, we outline the design of the model. The data of the model consist of $(y_{i}, x_{i1}, ......., x_{ik}), i = 1,......,n,$ for a binary response variable $y = \{0, 1\}$ with appropriately coded covariates $x_{1},......,x_{k}$. The probabilities of reactions of pedestrians $\pi_{i} = P(emotion_{i} = 1)$ are modeled by \[\pi_{i} = \frac{\exp(r_{i})}{1 + \exp(r_{i})}\]
with the linear predictor 
\begin{align*}
r_{i} &= \beta_{0} + \beta_{1}Interaction_{i1} +  \beta_{2}Gender_{i2} \\
&\quad+ \beta_{3}RobotFace_{i3} +  \beta_{4}Age_{i4} +  \beta_{5}Contact_{i5}
\end{align*} where $\beta$ are the coefficients of covariates.

\section{Results}
\label{sec:results}
\begin{figure*}[!t]
	\centering
	\includegraphics[width=0.8\textwidth]{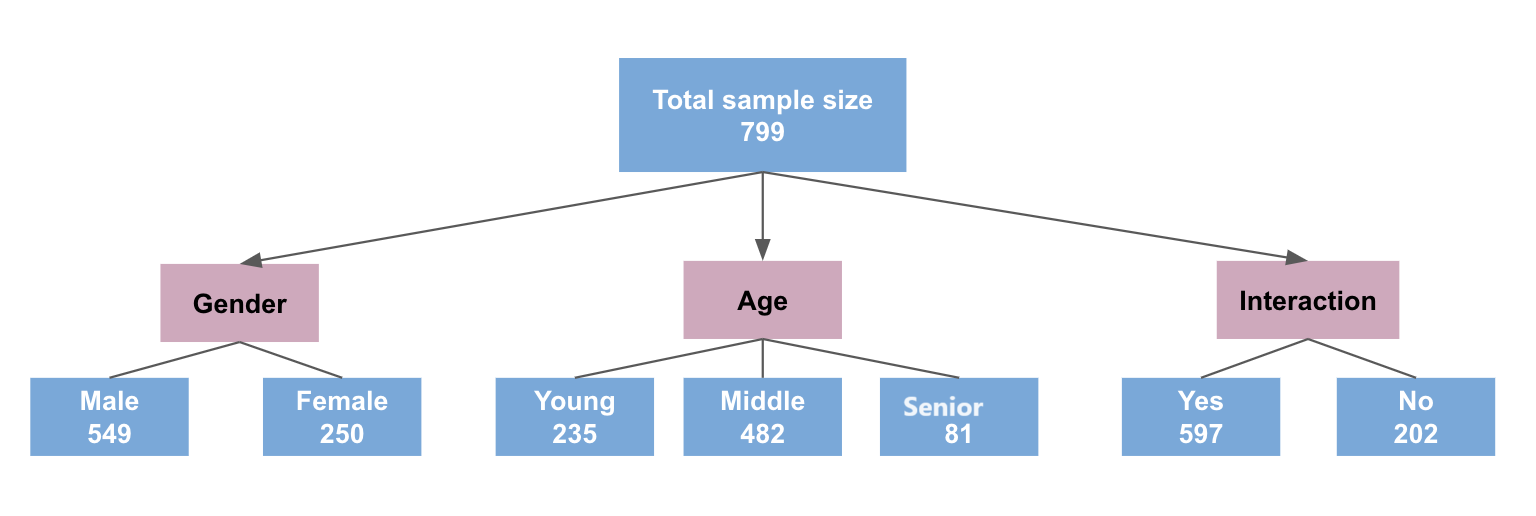}
	\caption{Sample sizes of covariates by count}
	\label{fig:sample-sizes-covariates}
\end{figure*}

A comprehensive sample of 799 individuals participated in our study. Fig.~\ref{fig:sample-sizes-covariates} shows the demographic breakdown, with 549 male and 250 female participants. Among these, the predominant demographic consisted of middle-aged pedestrians, encompassing 482 individuals, while a quarter of the participants, totaling 235, were identified as young individuals. A smaller segment of the sample, consisting of 81 individuals, represented the elderly demographic.

When examining interaction patterns with the LMD robot, out of the entire sample, 202 individuals did not engage in any interaction with the LMD robot, while a significant majority, comprising 597 individuals, actively interacted with the robot.

Throughout the experimental period, the behavior of 209 participants was examined in the presence of the LMD robot exhibiting an angry facial expression. Additionally, 190 pedestrians were observed under conditions where the robot displayed a happy facial expression, while a sizable group of 400 pedestrians exhibited no perceptible facial expressions during the course of the experiment.

Table \ref{tab:reactions} shows the summary of reactions to the happy versus angry faces facial expressions based on gender.

\begin{table}[ht]
\caption{Summary of Reactions to Facial Expressions Based on Gender (Total Sample)}
\centering
\begin{tabular}{lcc}
\toprule
\textbf{Variable} & \textbf{Men (\%)} & \textbf{Women (\%)} \\
\midrule
Engaged & 79.6 & 64 \\
\midrule
\multicolumn{3}{l}{\textbf{Happy Face}} \\
Engaged & 78.9 & 50.9 \\
\midrule
\multicolumn{3}{l}{\textbf{Angry Face}} \\
Engaged & 87.7 & 79.3 \\
\midrule
\multicolumn{3}{l}{\textbf{No Facial Expression}} \\
Engaged & 76.5 & 58.5 \\
\bottomrule
\end{tabular}
\label{tab:reactions}
\end{table}

\begin{figure}[!t]
	\centering
	\includegraphics[width=0.5\textwidth]{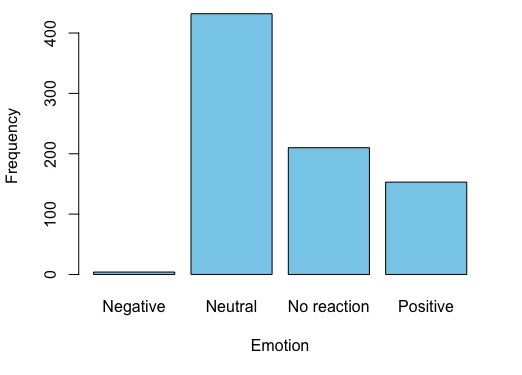}
	\caption{Distribution of the response variable pedestrians emotions}
	\label{fig:freq_ped_emotions}
\end{figure}

Despite conducting an ample number of experimental runs, the observed reactions from pedestrians were relatively divided. Specifically, a mere 4 individuals expressed negative reactions, while a substantial 153 individuals exhibited positive responses. Notably, a significant portion of the participants, accounting for more than half of the total sample (432 individuals), responded neutrally to the LMD robot. Furthermore, a quarter of the participants (210 individuals) did not exhibit any noticeable reaction. Fig.~\ref{fig:freq_ped_emotions} illustrates the distribution of participants' reactions.

\begin{figure}[!t]
	\centering
	\includegraphics[width=0.5\textwidth]{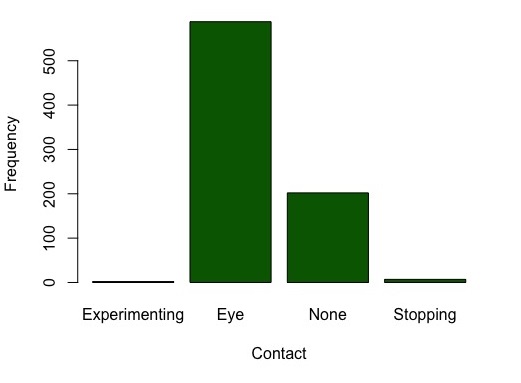}
	\caption{Distribution of the response variable pedestrians contact}
	\label{fig:freq_ped_contact}
\end{figure}

With regards to interpersonal engagement types, a predominant 98.5\% of interactions were characterized by the establishment of eye contact, involving a substantial sample of 588 cases. In contrast, the instances of experimenting interactions constituted a minimal 0.3\% of the total, with a count of 2. Likewise, stopping interactions comprised 1.2\% of the observed instances, with a count of 7. The distribution of these contacting types is visually represented in Fig.~\ref{fig:freq_ped_contact}.

As for the impressions gathered during the experiment, some pedestrians discretely attempted to film or take photos of the LMD robot. Many turned around after passing the robot for a second look. Generally, men stopped more often and observed the robot with fascination. People also tended to observe the robot while walking. Interestingly, the LMD robot with the angry face design resulted in several laughs from pedestrians. Alongside these observations, comments from pedestrians were collected separately and included: “Is this the new postman?”, “He (the LMD robot) is not very fast!”, “So cute!” (angry face design), and “Look, he'll soon replace the postman!”. Comments concerning the postal system probably derived from the design of the LMD robot, as the robot had several Austrian Post AG logos on prominent display.

\subsection{Results of Chi-square and Cramer's V tests}
To assess the associations among the variables of interest, we conducted a Chi-square test, at a 0.05 significance level. The null hypothesis for the Chi-square test assumes the absence of any relationships between the variables under consideration. Subsequently, Cramer's V test was employed to quantify the strength of relationships between these variables. The Cramer's V statistic ranges from 0 to 1, where a value of 0 signifies no relationship, while a value of 1 indicates a perfect relationship.

We focused on examining the relationships among the variables RobotFace, Reaction, Emotion, and Gender. The detailed outcomes of these statistical tests are presented in Table~\ref{tab:test-results}.

\begin{table}[ht]
    \caption{Chi-square and Cramer's V tests by variables}
    \centering
    \begin{tabular}{@{}lccc@{}}
        \toprule
        \textbf{Variables/Tests} & \textbf{Value (\%)} & \textbf{df (\%)} & \textbf{Sign. (\%)} \\
        \midrule
        \multicolumn{4}{@{}l}{\textbf{RobotFace vs Reaction vs Male}} \\
        Pearson Chi-Square  & 13.602 & 2 & 0.001 \\
        Cramer's V  & 0.130 & 2 & 0.001 \\
        \midrule
        \multicolumn{4}{@{}l}{\textbf{RobotFace vs Reaction vs Male}} \\
        Pearson Chi-Square  & 6.676 & 2 & 0.036 \\
        Cramer's V  & 0.110 &   & 0.036 \\
        \midrule
        \multicolumn{4}{@{}l}{\textbf{RobotFace vs Reaction vs Female}} \\
        Pearson Chi-Square  & 6.676 & 2 & 0.036 \\
        Cramer's V  & 0.110 &   & 0.036 \\
        \midrule
        \multicolumn{4}{@{}l}{\textbf{RobotFace vs Emotion}} \\
        Pearson Chi-Square  & 43.497 & 4 & $<$0.001 \\
        Cramer's V  & 0.191 &   & $<$0.001 \\
        \midrule
        \multicolumn{4}{@{}l}{\textbf{RobotFace vs Emotion vs Male}} \\
        Pearson Chi-Square  & 46.115 & 4 & $<$0.001 \\
        Cramer's V  & 0.230 &   & $<$0.001 \\
        \multicolumn{4}{@{}l}{\textbf{RobotFace vs Emotion vs Female}} \\
        Pearson Chi-Square  & 3.083 & 2 & 0.214 \\
        Cramer's V  & 0.139 &   & 0.214 \\
        \bottomrule
    \end{tabular}
    \label{tab:test-results}
\end{table}

It is evident that relationships exist among the variables of interest. However, the relationship between RobotFace, Emotion, and Female variables did not achieve statistical significance. The outcomes of the Cramer's V test indicate that all associations are statistically significant at the 0.05 level, with the exception of the relationship involving RobotFace, Emotion, and Female as revealed by the Chi-square test.

Despite statistical significance, the observed relationships across all variables exhibit a Cramer's V value less than 0.25. This suggests a weak degree of association among the variables under investigation.

\subsection{Model's result}
For the modeling process, cases where pedestrians exhibited no reaction were excluded. Based on the variable selection analysis, covariates including Age, Gender, and RobotFace were included in the model. Additionally, a minimal number of cases (4 pedestrians) initially categorized as displaying anger were reassigned to the neutral emotion category, reflecting the subjective nature of the interpretation of anger and the absence of explicitly happy reactions from the subjects. Table~\ref{tab:model-result} presents the results of the model.

\begin{table*}[htbp]
    \centering
    \caption{Results of the logit model}
    \begin{tabular}{lccccccc}
        \toprule
        \cmidrule{6-8} \cmidrule{5-7}
        & Variables & Est. Coefficients & Std. Error & z Value & $Pr(>|z|)$ \\
        \midrule
        Intercept & -0.823 & 0.254 & -3.233 &  0.001**  \\       
        Gender Women & -0.03 & 0.223 & -0.173 & 0.862  \\   
        Age Middle & 0.398 & 0.234 & 1.700 &  0.089  \\  
        Age Senior &  0.383 & 0.345 & 1.109 & 0.267  \\ 
        Faces Angry &  0.026 &  0.244 & 0.109 & 0.913  \\ 
        Faces Neutral & -1.267 &  0.251 & -5.047 & $4.49 \times 10^{-7}***$  \\  
        Null Deviance:  & 674.77  & on 588  degrees of freedom  \\ 
        Residual Deviance:  & 632.24  & on 583  degrees of freedom  \\ 
        AIC:  & 644.24  \\ 
    \end{tabular}
    \label{tab:model-result}
\end{table*}

As observed, gender does not have a noticeable effect, indicating that both females and males respond with similar emotions. Similarly, there is no notable distinction between middle-aged and senior pedestrians. However, young pedestrians tend to exhibit slightly happier emotions. 

Interestingly, there appears to be no significant difference between the effects of a happy or an angry facial expression of the LMD robot on pedestrians' reactions. 
Conversely, a neutral expression of the LMD robot seems to evoke a pronounced negative response from pedestrians, since the coefficient becomes negative when the robot's face is neutral. Therefore, the probability of a happy reaction decreases by approximately (1-0.2815571)*100\% = 72\% when the robot's face is neutral.

Given that the model categorizes reactions into only "Happy" and "Neutral" (with all other reactions presumed to fall under "Neutral"), we can infer the existence of other reactions besides happiness. However, the impact of a neutral face on the likelihood of a happy reaction is substantial.

The likelihood ratio test demonstrates that the model is highly statistically significant, with a p-value close to zero (4.598e-08). Consequently, the model rejects the null hypothesis and supports the alternative hypothesis, suggesting that pedestrians do indeed react to LMD robots featuring expressive facial features. The test results are displayed in 
Table~\ref{tab:likelihood-ration-test}.

\begin{table}[ht]
    \caption{Likelihood Ratio Test}
    \centering
    \begin{tabular}{lcccc}
        \toprule
        \textbf{\#Df} & \textbf{LogLik (\%)} & \textbf{Df} & \textbf{Chisq} & \textbf{Pr($>$Chisq)} \\
        \midrule
        6 & -316.12 & -- & -- & -- \\
        1 & -337.38 & -5 & 42.532 & $4.598 \times 10^{-8}***$ \\
        \bottomrule
    \end{tabular}
    \label{tab:likelihood-ration-test}
\end{table}

\begin{figure}[!t]
    \centering
    \includegraphics[width=0.5\textwidth]{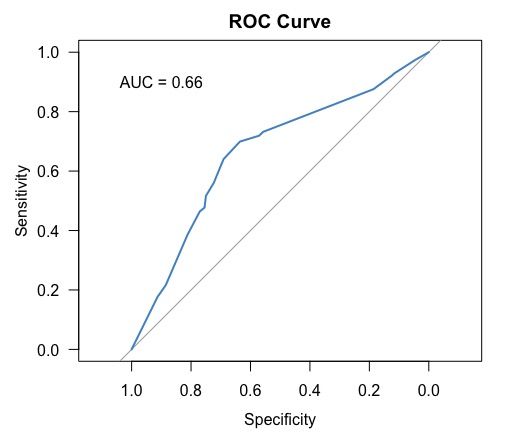}
    \caption{Model accuracy by ROC curve}
    \label{fig:ROC_curve}
\end{figure}

We assessed the model accuracy through a receiver operating characteristic (ROC) curve analysis. As shown in Figure \ref{fig:ROC_curve} the ROC curve indicates that the fitted model achieves moderate accuracy, as the area under the curve (AUC) is 0.66. However, enhancing the overall dataset size could potentially improve model performance by enabling it to glean more insights from the data.

\section{Conclusion and Future Work}
\label{sec:conclusion}
The objective of this paper was to study the interaction among facial expressions, and emotional responses in human-robot interactions in the context of last mile delivery.
The findings from this experiment indicate that even a rudimentary printout of a face attached to an LMD robot elicited more reactions compared to the LMD robot lacking any human-like features.

Incorporating even a simple facial design into these robots can significantly influence the experiences of pedestrians and recipients of packages and specific emotional expressions for the facial design can further impact the quantity and nature of reactions, potentially fostering greater acceptance of LMD robots.

Although the experiment involved nearly 800 pedestrians, the sample consisted mainly on university employees or students. To improve the generalizability of findings, multiple experiments across diverse demographics should be conducted in future research. 

While the emotional expressions of ``happy'' and ``angry'' provided initial insights into potential reactions toward LMD robots, incorporating a broader range of emotions into the facial design would deepen our understanding of their impact across different emotional contexts.

Finally, a logical progression in LMD robot design could involve creating robots capable of direct interaction with their surroundings, such as altering their facial expressions based on the situation.

\bibliographystyle{IEEEtran}
\bibliography{references}

\begin{thebibliography}{10}
\providecommand{\url}[1]{#1}
\csname url@samestyle\endcsname
\providecommand{\newblock}{\relax}
\providecommand{\bibinfo}[2]{#2}
\providecommand{\BIBentrySTDinterwordspacing}{\spaceskip=0pt\relax}
\providecommand{\BIBentryALTinterwordstretchfactor}{4}
\providecommand{\BIBentryALTinterwordspacing}{\spaceskip=\fontdimen2\font plus
\BIBentryALTinterwordstretchfactor\fontdimen3\font minus
  \fontdimen4\font\relax}
\providecommand{\BIBforeignlanguage}[2]{{%
\expandafter\ifx\csname l@#1\endcsname\relax
\typeout{** WARNING: IEEEtran.bst: No hyphenation pattern has been}%
\typeout{** loaded for the language `#1'. Using the pattern for}%
\typeout{** the default language instead.}%
\else
\language=\csname l@#1\endcsname
\fi
#2}}
\providecommand{\BIBdecl}{\relax}
\BIBdecl

\bibitem{willis2006first}
J.~Willis and A.~Todorov, ``First impressions: Making up your mind after a
  100-ms exposure to a face,'' \emph{Association for Psychological Science},
  vol.~17, no.~7, pp. 592--598, 2006.

\bibitem{zebrowitz1996social}
L.~A. Zebrowitz and J.~M. Montepare, ``Social psychological face perception:
  Why appearance matters,'' in \emph{Social Psychology}.\hskip 1em plus 0.5em
  minus 0.4em\relax Erlbaum, 1996, pp. 131--154.

\bibitem{hegel2014bedeutung}
F.~Hegel, ``Bedeutung von k{\"o}rper und form in der interaktion mit sozialen
  robotern,'' in \emph{Perulog - Freiburger Beitr{\"a}ge zur Kultur und
  Sozialforschung}, M.~Schetsche, Ed.\hskip 1em plus 0.5em minus 0.4em\relax
  Logos Verlag Berlin, 2014, vol.~7, pp. 84--103.

\bibitem{song2021facial}
Y.~Song, A.~Luximon, and Y.~Luximon, ``The effect of facial features on facial
  anthropomorphic trustworthiness in social robots,'' \emph{Applied
  Ergonomics}, vol.~94, p. 103420, 2021.

\bibitem{epley2008when}
N.~Epley, A.~Waytz, S.~Alkalis, and J.~T. Cacioppo, ``When we need a human:
  Motivational determants of anthropomorphism,'' \emph{Social Cognition},
  vol.~26, no.~2, pp. 143--155, 2008.

\bibitem{sagan1995demonhaunted}
C.~Sagan, \emph{The Demon-Haunted World – Science as a Candle in the
  Dark}.\hskip 1em plus 0.5em minus 0.4em\relax New York: Random House, 1995.

\bibitem{zlotowski2014anthropomorphism}
J.~Złotowski, D.~Proudfoot, K.~Yogeeswaran \emph{et~al.},
  ``Anthropomorphism,'' \emph{Opportunities and Challenges in Human–Robot
  Interaction}, vol.~7, pp. 347--360, 2014.

\bibitem{mori2012uncanny}
M.~Mori, K.~F. MacDorman, and N.~Kageki, ``The uncanny valley [from the
  field],'' \emph{IEEE Robotics \& Automation Magazine}, vol.~19, no.~2, pp.
  98--100, 2012.

\bibitem{mende2019use}
M.~A. Mende, M.~H. Fischer, and K.~K{\"u}hne, ``The use of social robots and
  the uncanny valley phenomenon,'' 2019.

\bibitem{aggarwal2007car}
P.~Aggarwal and A.~L. McGill, ``Is that car smiling at me?: Schema congruity as
  a basis for evaluating anthropomorphized products,'' \emph{Journal of
  Consumer Research}, vol.~34, pp. 468--479, 2007.

\bibitem{guthrie1993faces}
S.~E. Guthrie, \emph{Faces in the clouds: A new theory of religion}.\hskip 1em
  plus 0.5em minus 0.4em\relax Oxford University Press, 1993.

\bibitem{steffen2009brain}
A.~C. Steffen, B.~Rockstroh, and B.~Jansma, ``Brain evoked potentials reflect
  how emotional faces influence our decision making,'' \emph{Journal of
  Neuroscience, Psychology, and Economics}, vol.~2, no.~1, pp. 32--40, 2009.

\bibitem{novotny2023ros}
G.~Novotny, W.~Morales-Alvarez, N.~Smirnov, and C.~Olaverri~Monreal,
  ``Development of a ros-based architecture for intelligent autonomous on
  demand last mile delivery,'' 2023.

\bibitem{jku}
JKU, ``Jku campus,'' \url{https://www.jku.at/campus/der-jku-campus/gebaeude/},
  2024.

\bibitem{darkspeeds}
Darkspeeds, ``Facial expressions buddy sheet for comics/cartoons,''
  \url{https://www.deviantart.com/darkspeeds/art/Facial-Expressions-Buddy-Sheet-for-comics-cartoons-359562676},
  accessed: [Insert Access Date].

\end{thebibliography}

\end{document}